\documentclass[copyright,noncommercial]{eptcs}
\pdfoutput=1
\providecommand{\volume}{5}
\providecommand{\anno}{2009}
\providecommand{\firstpage}{41}
\providecommand{\eid}{4}
\providecommand{\event}{Y. Deville and C. Solnon (Eds): Sixth International Workshop\\
 on Local Search Techniques in Constraint Satisfaction (LSCS'09).}
\usepackage{breakurl}        
\usepackage[lined,algonl,boxed]{algorithm2e} 

\title{On Improving Local Search for Unsatisfiability}

\author{David Pereira \quad In\^es Lynce
\institute{IST/INESC-ID, Technical University of Lisbon, Portugal}
\email{{david,ines}@sat.inesc-id.pt} \\
\and 
\and Steven Prestwich
\institute{Cork Constraint Computation Centre, Dept of Computer Science, \\
University College, Cork, Ireland}
\email{s.prestwich@cs.ucc.ie}
}
\def\titlerunning{On Improving Local Search for Unsatisfiability}
\def\authorrunning{David Pereira {\em et al.}}

\newtheorem{example}{Example}

\begin{document}
\maketitle

\begin{abstract}
Stochastic local search (SLS) has been an active field of research in
the last few years, with new techniques and procedures being developed
at an astonishing rate. SLS has been traditionally associated with
satisfiability solving, that is, finding a solution for a given
problem instance, as their intrinsic nature does not address
unsatisfiable problems. Unsatisfiable instances were therefore
commonly solved using backtrack search solvers. For this reason, in
the late 90s Selman, Kautz and McAllester proposed a challenge to use
local search instead to prove unsatisfiability.  More recently, two
SLS solvers -- {\sc Ranger} and {\sc Gunsat} -- have been developed,
which are able to prove unsatisfiability albeit being SLS solvers. In
this paper, we first compare {\sc Ranger} with {\sc Gunsat} and then
propose to improve {\sc Ranger} performance using some of {\sc
Gunsat}'s techniques, namely unit propagation look-ahead and extended
resolution.
\end{abstract}

\section{Introduction}

Selman, Kautz and McAllester posed an intriguing challenge in 1997 to
use local search to prove unsatisfiability rather than satisfiability
\cite {challenge97}.  In 2006 and 2007, two different approaches were
proposed in response to that challenge \cite {ranger,gunsat}. These
two algorithms -- {\sc Ranger}~\cite{ranger} and {\sc
Gunsat}~\cite{gunsat} -- use local search to prove unsatisfiability
instead of satisfiability, still being incomplete algorithms. They
can, however, prove that a formula is satisfiable under certain
conditions.

Previous work has addressed the use of hybrid algorithms combining
clause learning with local search to prove either satisfiability or
unsatisfiability~\cite{richards-jar00,ruml-aaai04,zhang-lpar05} at the
cost of having worst-case exponential space complexity. Alternative
approaches combined unit clause elimination and local
search~\cite{hirsh-cp01}.

We should note, however, that the algorithms just mentioned clearly
differ from {\sc Ranger} and {\sc Gunsat}, which were built with the
main goal of proving unsatisfiability by (directly or indirectly)
deriving the empty clause. The main idea is to apply a sequence of
resolution steps and other satisfiability preserving techniques to
conclude that the formula has no solution.

Although based on the same ideas, {\sc Ranger} and {\sc Gunsat}
differ.  {\sc Ranger} is a lightweight algorithm that performs many
moves per second, while {\sc Gunsat} applies more powerful reasoning
techniques.  This paper presents for the first time a detailed
comparison of these two algorithms and further integrates some of {\sc
Gunsat} techniques into {\sc Ranger}.

This paper is organised as follows. The next section provides the
required background.  Section~\ref{ranger+gunsat} describes and
compares {\sc Ranger} and {\sc Gunsat}. Afterwards, we describe the
integration of {\sc Gunsat}'s techniques into {\sc
Ranger}. Section~\ref{sec:results} provides the experimental
evaluation of the new techniques. Finally, the paper concludes.

\section {Background}

We assume the basic definitions in the context of propositional
satisfiability. A propositional formula $\varphi$ is a conjunction of
clauses, where a clause $c$ is a disjunction of literals and a literal
$l$ is either a variable or its negation, being either a positive or a
negative literal, respectively. Each variable $\nu$ can be assigned a
truth value ({\tt true} or {\tt false}, also often referred to as 1 or
0, respectively).  A positive literal $l$ is said to be satisfied
(unsatisfied) if the corresponding variable is assigned value {\tt
true} ({\tt false}). A negative literal $l$ is said to be satisfied
(unsatisfied) if the corresponding variable is assigned value {\tt
false} ({\tt true}). A clause is said to be satisfied if at least one
of its literals is satisfied, unsatisfied if all of its literals are
unsatisfied and unresolved otherwise. Unresolved clauses with only one
unassigned literal are said to be unit. A formula is satisfied if all
of its clauses are satisfied.
The propositional satisfiability (SAT) problem consists in deciding
whether there exists a truth assignment to the variables such that the
formula becomes satisfiable. Such an assignment is called a solution.

Two of the most well-known complete algorithms for SAT, which both
still inspire current state of the art algorithms, are the
Davis-Putnam (DP)~\cite{davis-jacm60} and the Davis-Logemann-Loveland
(DLL)~\cite{davis-cacm62} algorithms. The first one is based on the
resolution principle~\cite{resolution}, whereas the second one is
based on backtrack search. An important feature on a backtrack search
algorithm for SAT is the use of unit propagation. Clearly, a unit
clause has only one way to be satisfied, which implies satisfying its
unassigned literal. This rule may be iteratively applied until no unit
clauses remain in the formula.

Local search algorithms, in contrast, are incomplete as they are not
able to prove unsatisfiability: either they find a solution or the
answer is {\em unknown}, which means that either the formula has no
solution or the solver was not able to find a solution in the allowed
CPU time.

Local search algorithms start at some location in the given problem's
search space and then move from the start location to a neighbouring
location in the search space which is determined by a decision based
on local knowledge only. These local search algorithms are typically
incomplete, that is, there is no guarantee that an existing solution
will be found, and if no solution exists that fact can never be
determined with certainty. Furthermore, these search methods can visit
the same location in the search space more than once, and they can
become trapped in a small number of locations from which they cannot
escape: these are called {\em local minima}, which will be discussed
shortly, and require special {\em escape strategies}.

Many widely known and high-performance local search algorithms make
use of randomized choices when generating and/or selecting candidate
solutions for a given problem \cite {sls}. These algorithms are called
{\em stochastic local search} (SLS) algorithms, and they belong to the
most powerful methods for practically solving large and hard
satisfiable instances of SAT, and outperform the best systematic
search methods on a number of domains. In SLS algorithms, the initial
position in the search space is chosen randomly, as are the decisions
to move from a position to another.
Two early but influential stepping stones in SLS algorithm development
are GSAT \cite {gsat} and WalkSAT \cite {walksat}.

\section {Local Search for Unsatisfiability}
\label{ranger+gunsat}

{\sc Ranger} and {\sc Gunsat} are two stochastic local search
algorithms which resemble the skeleton of SLS algorithms, but on the
other hand aim at deriving the empty clause to prove unsatisfiability.

\subsection {{\sc Ranger}}

{\sc Ranger} \cite {ranger} stands for RANdomized GEneral Resolution
and was presented in 2006 as the first SLS algorithm that can prove
unsatisfiability rather than satisfiability. It explores a space of
multisets of resolvents using general resolution and aims at deriving
the empty clause non-systematically but greedily, thus proving
unsatisfiability. {\sc Ranger} will eventually refute any
unsatisfiable instance while using only bounded memory.

A theoretical result behind the exploration of local search on
multisets of resolvents can be found in \cite
{multisets_resolvents}. The authors show that the space needed 
for a resolution proof is no more than $n+1$ resolvents, where $n$ 
is the number of variables in the problem. 

Given an unsatisfiable SAT 
formula $\varphi$ with {\em n} variables and {\em m} clauses, 
a general resolution refutation can be represented by a series of formulae
$\varphi_1$,...,$\varphi_s$ where $\varphi_1$ consists of some or all
of the clauses in $\varphi$, and $\varphi_s$ contains the empty
clause. Each $\varphi_i$ is obtained from $\varphi_{i-1}$ by
(optionally) deleting some clauses in $\varphi_{i-1}$, adding the
resolvent of two clauses in $\varphi_{i-1}$, and (optionally) adding
clauses from $\varphi$. The {\em space} of a proof is defined as the
minimum {\em k} such that each $\varphi_i$ contains no more than {\em
k} clauses.

Intuitively, each $\varphi_i$ represents the set of {\em active}
clauses at step {\em i} of the proof. Inactive clauses are not
required for future resolutions, and after they have been used as
needed they can be deleted (for example clauses that are subsumed by
others).

The {\em width} of a proof is the length (in literals) of the largest
clause in the proof. Any non-tautologous clause must have length no
greater than {\em n}, so this is a trivial upper bound for the {\em
width} used in {\sc Ranger}. However, in practice, it may succeed even
if the resolvent length is restricted to a smaller value, which will
save memory on large problems.
Each $\varphi_i$ will be of the same constant size, and derived from
$\varphi_{i-1}$ by the application of resolution or the replacement of
a clause by one taken from $\varphi$.

The architecture of {\sc Ranger} is shown in algorithm \ref
{alg:ranger}. It has six parameters: the formula $\varphi$, three
probabilities $p_i$, $p_t$ and $p_g$, the width $w$ and the size $k$
of the formula $\varphi_i$.

\incmargin{1em}
\restylealgo{boxed}
\begin{algorithm}[t]
\caption{The {\sc Ranger} algorithm}
\label{alg:ranger}
\BlankLine
	\SetKwInOut{Input}{Input}
	\SetKwInOut{Output}{Output}
	\SetKwBlock{With}{with}{end}
	\Input {formula $\varphi$ in CNF, $p_i$, $p_t$, $p_g$, $w$, $k$}
	\Output {{\sc Unsatisfiable} {\bf or} {\sc Unknown}}
	\For {$try$ = 1 to $maxTries$} {
		$i = 1$ and $\varphi_1 = $\{any $k$ clauses from $\varphi$\} \\
		\For {$step$ = 1 to $maxSteps$} {
			\If {$\varphi_i$ contains the empty clause} {
				\Return {\sc Unsatisfiable} }
			\With (probability $p_i$) {
				replace a random $\varphi_i$ clause by a random $\varphi$ clause}
			\Other {resolve random $\varphi_i$ clauses $c$,$c'$ giving $r$ \\
			\If {$r$ is non-tautologous {\bf and} $|r| \leq w$} {
				\With (probability $p_g$) {
					\If {$|r| \leq max(|c|,|c'|)$} {
						replace the longer of $c$,$c'$ by $r$}}
				\Other {replace a random $\varphi_i$ clause by $r$}}}
			\With (probability $p_t$) {
				apply any satisfiability-preserving transformation to $\varphi$,$\varphi_i$ }
			$i = i+1$ and $\varphi_{i+1} = $\{the new formula\} }}
	\Return {\sc Unknown}
\end{algorithm}%

The {\sc Ranger} algorithm begins by choosing any $k$ clauses from the
formula $\varphi$ into $\varphi_1$. It then performs $i$ iterations,
either replacing a $\varphi_i$ clause with a $\varphi$ clause (with
probability $p_i$) or resolving two $\varphi_i$ clauses and placing
the result $r$ into $\varphi_i$. In the latter case, if $r$ is a
tautology or contains more than $w$ literals then it is discarded and
$\varphi_{i+1} = \varphi_i$.  Otherwise a $\varphi_i$ clause must be
removed to make room for $r$: either (with probability $p_g$) the
removed clause is the longer of the two parents of $r$ or it is
randomly chosen. In the former case, if $r$ is longer than the parent
then $r$ is discarded and $\varphi_{i+1} = \varphi_i$. With
probability $p_t$ any satisfiability-preserving transformation may be
applied to $\varphi$, $\varphi_i$ or both. One can apply subsumption
and the pure literal rule in several ways as satisfiability-preserving
transformations. If the empty clause has been derived then the
algorithm returns {\sc Unsatisfiable}, otherwise it may not
terminate. A time-out condition may be added to restrict the CPU time
that the algorithm is allowed to run.

In this algorithm the goal is to derive the empty clause, and as such
$\varphi_i$ must contain some small clauses. This is controlled by the
level of greediness (probability $p_g$). A greedy local move is one
that does not increase the number of literals in $\varphi_i$. So,
increasing $p_g$ will increase the greediness of the search, reducing
the proliferation of large resolvents.

{\sc Ranger} has a useful convergence property: for any unsatisfiable
SAT problem with {\em n} variables and {\em m} clauses, it finds a
refutation if $p_i > 0$, $p_i, p_t, p_g < 1$, $w = n$ and $k \geq
n+1$. For a proof, see \cite {ranger}.  The space complexity of {\sc
Ranger} is $O \left( n + m + kw \right) $. To guarantee convergence,
it requires $w = n$ and $k \geq n + 1$ so the space complexity becomes
at least $O \left( m + n^2 \right)$. In practice, it may require {\em
k} to be several times larger, but a smaller value of {\em w} is
usually sufficient.

It should be noted that {\sc Ranger} performs very poorly on
unsatisfiable random 3-SAT problems. This is an interesting asymmetry,
given that local search performs well on satisfiable random
problems. This may be because such refutations are almost certainly
exponentially long \cite {hard_resolution_examples}.

\subsection {{\sc Gunsat}}

{\sc Gunsat} \cite {gunsat} proposes to make a greedy walk through the
resolution search space in which, at each iteration of the algorithm,
it tries to compute a better neighbouring set of clauses, i.e. a set
of clauses similar to the previous one, differing from the previous
one by at most two clauses: one added by resolution and one that may
have been removed.
Intuitively, it will add new clauses and remove existing ones to the
formula, trying to derive the empty clause by using the resolution
rule.

{\sc Gunsat} is depicted in algorithm \ref {alg:gunsat}.  It either
proves that a problem instance is {\sc Unsatisfiable} or, if it does
not derive an empty clause within {\em maxTries} then it returns {\sc
Unknown}.  Also, if {\sc Gunsat} fails to derive the empty clause
after $maxSteps$ a restart is performed. By then all clauses, except
vital and binary clauses, are removed.  {\em Vital clauses} are
initial clauses, or any clause that previously subsumed another vital
clause. They ensure that the unsatisfiability of the formula is
preserved.

\incmargin{1em}
\restylealgo{boxed}
\begin{algorithm}[t]
\caption{The {\sc Gunsat} algorithm}
\label{alg:gunsat}
\BlankLine
	\SetKwInOut{Input}{Input}
	\SetKwInOut{Output}{Output}
	\Input {formula $\varphi$ in CNF}
	\Output {{\sc Unsatisfiable} {\bf or} {\sc Unknown}}
	\For {$try$ = 1 to $maxTries$} {
		\For {$step$ = 1 to $maxSteps$} {
			\If {2-saturation($\varphi$) returns {\sc Unsatisfiable}} {
				\Return {\sc Unsatisfiable}}
			\If {$|\varphi| > MaxSize$} {
				{\em remove-one-clause}($\varphi$) }
			{\em add-one-clause}($\varphi$) \\
			{\em add-extended-variables}($\varphi$) \\
			{\em simplify-look-ahead}($\varphi$) }
		replace $\varphi$ by all its {\em vital} and {\em binary} clauses }
	\Return {\sc Unknown}
\end{algorithm}%

{\sc Gunsat} operates on an initial formula $\varphi$ through a few
operations: {\em 2-saturation, remove-one-clause, add-one-clause,
add-extended-variables} and {\em simplify-look-ahead}.  Some of these
operations depend on a powerful scoring scheme. A score is given to
all pairs of literals based on their frequency appearance in the
formula.  Let us consider a clause $c_i$ of length $n_i$. Each pair
$\left( l_1, l_2 \right)$ appearing in $c_i$ is credited a weight of
$w_2 \left( n_i \right) = \frac{2^{n-1-n_1}}{n_i . \left(n_i - 1
\right)}$.  The score of a pair of literals $\left( l_1, l_2 \right)$
is defined as the sum of its weights in all clauses and noted $S
\left( l_1, l_2 \right)$.  The score $S \left( c \right)$ of a clause
{\em c} is the sum of the scores of all the pairs of literals it
contains.

The 2-saturation step ensures that, each time a new binary clause is
added to $\varphi$, all resolution operations between the set of
binary clauses are performed to saturate $\varphi$. In order to
exploit their full power, an equivalency literal search is
performed. While performing the binary clause saturation, the
algorithm may find new unit clauses (note that in a resolution step
between two binary clauses the resolvent can have either one or two
literals). The literal $l$ of the unit clause is then propagated in
the whole formula by unit propagation. An inconsistency may be
identified at this step and the algorithm returns {\sc Unsatisfiable},
proving the unsatisfiability of the formula. Refer to \cite
{bin_saturation} for the use of binary clause saturation for
preprocessing purposes.

In addition, if the size of the current formula is greater than a
fixed $MaxSize$ then a non-vital clause is removed by {\em
remove-one-clause}. In each iteration there is also a call to {\em
add-one-clause}, which adds one clause to the current formula
according to the scoring scheme. Both {\em add-extended-variables} and
{\em simplify-look-ahead} use reasoning mechanisms (extended
resolution in the former, unit propagation look-ahead in the latter)
to improve the chances of deriving an empty clause in the next
iteration.  {\em Add-extended-variables} adds the three clauses
generated through extended resolution to the formula. {\em
Simplify-look-ahead} applies unit propagation look-ahead to the
formula~\cite{lookahead}, which may eventually conclude the formula is
unsatisfiable.

Extended resolution is applied when the algorithm has tried to
increase the score of a given pair of literals too many times without
success, and it uses extended resolution to artificially increase that
score. The application of the extended rule implies adding a new
variable $e$ and three clauses to the formula.  In practice, $e
\Leftrightarrow l_1 \vee l_2$ is encoded by the three clauses $(\neg e
\vee l_1 \vee l_2)$, $(e \vee \neg l_1)$ and $(e \vee \neg l_2)$.

Look-ahead techniques are used to detect equivalences between literals
until an inconsistency is found. {\sc Gunsat} uses look-ahead unit
propagation on pairs of literals, such that the four possible pairs of
values are propagated in $\varphi$, potentially implying more
propagations.

\subsection {{\sc Ranger} vs. {\sc Gunsat}}

There are some few important differences between these two local
search algorithms for proving unsatisfiability. {\sc Ranger} generates
a large number of the shortest possible clauses as fast as possible,
using unintelligent local moves, whereas {\sc Gunsat} takes longer to
make more intelligent moves based on a more complex objective
function. {\sc Gunsat} also uses higher reasoning techniques like
extended resolution and unit propagation look-ahead ({\sc Ranger} uses
only general resolution).  Also, unlike {\sc Gunsat}, {\sc Ranger}
uses a mechanism to ensure bounded memory.

\section {Improving {\sc Ranger}}

This section describes the implementation of unit propagation
look-ahead and extended resolution in {\sc Ranger}. We should note
that before starting to implement new features into the solver we made
a series of modifications to the original tool in terms of data
structures to better accommodate our needs. For the new techniques, it
is required to have a complete knowledge of a clause status, namely to
identify whether it is satisfied, unsatisfied or unresolved, in which
case is important to distinguish unit clauses.

\subsection {Implementing Unit Propagation Look-Ahead}

The solver {\sc Gunsat} successfully uses a method dubbed unit
propagation look-ahead \cite{gunsat} to improve its basic
algorithm. The literals of the formula under consideration are
extensively checked to see if there are any conflicts arising from
hypothetical assignments.

{\sc Gunsat} implements a version of the unit propagation look-ahead
which uses two variables. These two variables are then assigned a
value such that the four possible combinations are covered. If $\nu_1$
and $\nu_2$ are our variables, then the four possible combinations
are: (1) $\nu_1 = 0$ and $\nu_2 = 0$, (2) $\nu_1 = 0$ and $\nu_2 = 1$,
(3) $\nu_1 = 1$ and $\nu_2 = 0$, and (4) $\nu_1 = 1$ and $\nu_2 = 1$.
        
Given an assignment to a pair of variables, a {\em conflict} is
identified when one of the clauses becomes unsatisfied as a result of
unit propagation. For each iteration of this look-ahead method, i.e.,
for each combination of variable assignments, we store the value of
each variable in the formula, only if that variable is forced to be
assigned as a result of unit propagation (note that these assignments
are only temporary, done for each iteration of the look-ahead and
stored only for the duration of the look-ahead for the two variables).
But in {\sc Ranger} we have further extended this technique.

Let us consider the assignments which were implied by unit propagation
after assigning a pair of variables. We may further consider {\em intersections}
of implications when taking into account different assignments made to
the variables in that pair. We should now focus on the number of conflicts 
after the application of the unit propagation look-ahead technique after 
assigning a pair of variables. 
In the worst case, we may end up with four conflicts, one
for each of the four different assignments for a given pair of variables. 
(Note that the method to be
applied resembles the St\"almark's method \cite {stalmark} and has
been applied to CNF formulas in the past~\cite{sat-ictai03}.)
Overall, we have five possible scenarios:

\begin {itemize}
        \item {\bf Zero conflicts}: If there are no conflicts,
        we will consider all four combinations when computing
        the intersection. If a variable is assigned the same
        value through all combinations, then that value will be 
        assigned and the unit clause rule will be applied.
        \item {\bf One conflict}: 
        \begin {itemize}
                \item The intersections will be
        calculated, but now only considering the combinations that
        did not yield a conflict (three in this case). Again, if
        a variable is assigned the same value through the three 
        combinations, then that value will be 
        assigned and the unit clause rule will be applied.
                \item A binary clause is added to
        the formula: this clause results from the negation of 
        the assignments that yield a conflict.
        \end {itemize}
        \item {\bf Two conflicts}:
        \begin {itemize}
                \item As above, the intersections will be
        calculated, but only considering the combinations that
        did not yield a conflict (two in this case). This may result
        in new unit clauses.
                \item Two binary clauses are added to the formula, 
        resulting each one from the negation of the assignments
        that yield a conflict. There is a special case where
        only one unit clause is added, which happens when a variable 
        assignment is repeated in both conflicts (the two binary
        clauses are resolved to generate the unit clause).
        \end {itemize}
        
        \item {\bf Three conflicts}:
        \begin {itemize}
                \item The values that were assigned in the only
        combination that did not yield a conflict will be
        propagated as a result of the two unit clauses being added, 
        each one with each the variable assignment that did not yield 
        a conflict.
        \end {itemize}
        \item {\bf Four conflicts}: The formula yields a conflict
        for all combinations, which means that the formula is 
        unsatisfiable.
\end {itemize}

\begin {example}
To illustrate the look-ahead behaviour of {\sc Gunsat} suppose we have
the following formula $\varphi$ with the clauses:

\begin{center}
$c_1 = \left(\nu_1\vee\nu_2\vee\nu_3\right)$
\\$c_2 = \left(\nu_1\vee \neg\nu_2\vee\nu_3\right)$ 
\\$c_3 = \left( \neg\nu_1\vee\nu_3\right)$
\\$c_4 = \left( \nu_3 \vee \neg \nu_4 \right)$
\end{center}

Clearly, for any of the possible assignments to $\nu_1$ and $\nu_2$
the value of $\nu_3$ must be 1. Thus, we have that $\nu_3$ must be
assigned value 1, regardless of the assignments made to other
variables, and the clause $c_5 = (\nu_3)$ can be added to the formula
as a unit clause. We can then perform unit propagation.
\end {example}

\begin {example}
The unit propagation look-ahead can also be used to derive an empty
clause.  Suppose we have the following formula $\varphi$:

\begin{center}
$c_1 = \left(\nu_1\vee \nu_2\right)$ \\
$c_2 = \left(\neg\nu_1\vee \nu_2\right)$ \\
$c_3 = \left(\neg\nu_2\vee \nu_3\right)$ \\
$c_4 = \left(\neg\nu_2\vee \neg\nu_3\right)$
\end{center}

If we consider $\nu_2 = 1$, then either clause $c_3$ or $c_4$ becomes
unsatisfied. On the other hand, if we consider $\nu_2 = 0$, then
either clause $c_1$ or $c_2$ becomes unsatisfied.  Therefore, there is
no assignments to variable $\nu_1$ to make the formula satisfied and
as such we conclude that it is unsatisfiable.  (In this case, there
was no need to consider pairs of assignments to reach such a
conclusion.)
\end {example}

One of the objectives of this paper is to successfully integrate
features of the {\sc Gunsat} algorithm into the {\sc Ranger}
algorithm. We do not want, however, to modify the most important
properties of the original {\sc Ranger}, nor alter its flow. The unit
propagation look-ahead was, thus, added to the step of
satisfiability-preserving transformations.  The probability, $P_t$, to
execute these transformations is 90\%, and like the other
transformations, unit propagation will be executed, on average, in
90\% of the iterations of the algorithm.

This procedure has been divided into two parts: unit propagation
look-ahead with pairs of literals and unit propagation look-ahead with
only one literal. The first one is only executed once, the first time
that satisfiability-transformations are executed, due to the overhead
it has on the performance of the solver.
The second part of this procedure, which is less expensive, is
executed in every satisfiability-preserving transformation.

Finally, note that these methods can prove the unsatisfiability of a
formula themselves: if all possible assignments to a pair of variables result in
conflicts then the formula is unsatisfiable; likewise, if for both
possible assignments to a variable a conflict is detected, the
algorithm also returns {\sc Unsatisfiable}. It is also possible to
find a solution during this step, though this occurs less frequently.

\subsection {Implementing Extended Resolution}

The way extended resolution is used in {\sc Gunsat} is intrinsically
related to the algorithm itself, built to take advantage of its
scoring scheme. Note that extended resolution is only used when the
algorithm has tried to increase the score of a pair of literals too
many times without success. It is very different from the way {\sc
Ranger} operates, where no scoring scheme for literals is used. Thus,
we had to add {\sc Gunsat}'s scoring scheme to {\sc Ranger} to
implement extended resolution in the same way it was successfully used
in {\sc Gunsat}.

But there is a problem with this approach. {\sc Gunsat}'s scoring
system for pairs of literals is part of the main heuristic of the
solver. It was developed to be the backbone of the algorithm and to be
a highly refined heuristic of scoring for choosing the best literals
and clauses to resolve. Methods like extended resolution, unit
propagation look-ahead or binary clause saturation are only meant to
improve this heuristic.

In the previous section, we described the way in which unit
propagation look-ahead, a method also used in {\sc Gunsat} and which
proved to yield successful results, was added to the original {\sc
Ranger}. But this method was not intrinsically linked to the base of
{\sc Gunsat}, as extended resolution is. One could simply add it as a
preprocessing technique, or run it as a satisfiability preserving
transformation to {\sc Ranger} without loss of identity.

Even though adding extended resolution to {\sc Ranger} seemed to be
neither more efficient nor an easy task, and although it did not
promise to integrate well with the already implemented algorithm and
methods, we tried to integrate it with {\sc Ranger} and to improve its
performance on unsatisfiable instances.  As said above, {\sc Ranger}
does not have a scoring scheme for literals like {\sc Gunsat} does, so
that extended resolution could be applied directly to that scheme and
be integrated seamlessly in the algorithm. Instead, we chose to adapt
the scoring method of {\sc Gunsat} to {\sc Ranger} and thus apply the
extended resolution in the same way {\sc Gunsat} does.

Extended resolution is executed only during the
satisfiability-preserving transformations phase of the {\sc Ranger}
algorithm, in the same way as unit propagation look-ahead with one
variable is, and for the same reason: we did not want to alter {\sc
Ranger}'s base concept and program flow. Furthermore, besides the
probability $P_t$ of this phase of the algorithm, we inserted another
probability $P_{er}$, and the steps of extended resolution will only
be executed according to this probability. We had some trouble finding
an appropriate number for this probability, mostly because, in the
{\sc Gunsat} paper \cite {gunsat} the authors write that extended
resolution is used after the algorithm has tried to increase the score
of a given pair of literals ``too many times'', but do not provide
details. We chose $P_{er} = 5\%$ which means it will be executed in
about 4.5\% of the iterations (calculated by multiplying the 90\%
chance that the satisfiability-preserving transformations phase will
be run and the 5\% chance that extended resolution will be executed
within that phase).

At the start of each phase of extended resolution, our algorithm will
compute the scores for all the pairs of literals in the working
formula, in the same way that {\sc Gunsat} calculates its scores: each
pair of literals ($l_1,l_2$) appearing in the formula is credited a
score computed by adding the score of each literal in each clause
$c_i$ according to the following formula: $\frac{2^{n-1-n_1}}{n_i
. \left(n_i - 1 \right)}$. We compute the score of each variable in
each clause and add them for each pair of literals.  The score of a
clause is then computed by summing the scores of all the pairs of
literals it contains. Finally, we calculate the score of each
quadruplet by adding the sum of the squares of the scores of its pairs
of literals.

After the scores have been computed, we continue to follow {\sc
Gunsat}'s heuristic to improve the score of a pair of literals (note
that extended resolution will only be applied if we cannot improve the
score of a pair of literals after too many times). The best scored
quadruplet in the formula is computed and found, containing the
literals $l_1$ and $l_2$. We then try to find a new clause with both
$l_1$ and $l_2$ by searching the working formula for two clauses: one
containing $l_1$ and a pivot literal $p$, and the other containing
$l_2$ and the complement of the pivot literal, $\neg p$, such that the
resolution rule can be performed and the needed clause with $l_1$ and
$l_2$ is generated. We only try to generate the new clause $c$ from
the two clauses having the lowest scores. Because of this restriction,
it is not always possible to generate a new clause according to the
specified conditions (the cause could simply be that the new clause
may be subsumed by an already existing clause, or it may be a
tautology). If the score of one pair of the highest scored quadruplet
cannot be improved, the other scores of the same quadruplet are
iteratively tried. If no pairs of literals in this quadruplet can be
improved, the second best scored quadruplet is tried and so on.

Finally, the algorithm checks, for all pairs of literals, whether
their score has been increased too many times without any success or
not. We set this value to 20, i.e., we consider the score of a pair of
literals to be increased too many times when that value reaches 20. If
this occurs for literals $l_1$ and $l_2$ then three new clauses will
be added along with a new variable $e$:
\begin{center}
$c_1  = \left( \neg e \vee l_1 \vee l_2 \right)$
\\$c_2  = \left( e \vee \neg l_1 \right)$
\\$c_3  = \left( e \vee \neg l_2 \right)$
\end{center}

As we can see, there is a lot of implementation work needed to add
this method to {\sc Ranger}, especially if we apply it in the same way
{\sc Gunsat} does. Computing the scores and looking for the new clause
is very demanding for the algorithm, and we are adding an additional
layer of complexity to {\sc Ranger}, whereas in {\sc Gunsat} this
computation was already part of the algorithm itself.

\section{Experimental Evaluation}
\label{sec:results}

This section illustrates the behaviour of {\sc Ranger} on a set of
problem instances.  The instances consist of two
benchmarks~\footnote{Available from {\tt
www.cs.ubc.ca/\~{}hoos/SATLIB/benchm.html}.}:

\begin {itemize}
\item The {\tt aim} benchmark instances are all generated with a
particular random 3-SAT instance generator~\cite{aim96}. Its primary
role is to provide instances that the conventional random generation
cannot generate. The generator runs in a randomized fashion, so that
it is essentially different from those generated deterministically, or
those translated from other problems such as graph colouring.  We
utilized three sets of unsatisfiable {\tt aim} problem instances. Each
set has either 50, 100 or 200 variables, and within each set we have
two groups: four instances where the ratio clause/variable is 1.6, and
another group of four instances for which the ratio is 2.
\item The {\tt uuf50-218} benchmark consists of unsatisfiable uniform random 3-SAT
instances. For an instance with $n$ variables and $k$ clauses, each of
the $k$ clauses has three literals which are randomly picked from the
$2n$ possible literals (the $n$ variables and their negations) such
that each possible literal is selected with the same
probability. Clauses are discarded either if they contain repeated
literals or a literal and its negation (i.e. tautologous clauses). We
considered the data set of 100 unsatisfiable instances with 50
variables and 218 clauses, dubbed data set {\tt uuf50-218}.
\end {itemize}

Table~\ref{instances} illustrates the characteristics of each set of
problem instances, providing for each of them the number of instances,
variables and clauses.

\begin{table}[t]
\begin{center}
\begin{tabular}{|c|ccc|}
\hline
Instance&\#instances&\#variables&\#clauses\\
\hline
$aim-50-no-1\_6$&4&50&80\\
$aim-50-no-2\_0$&4&50&100\\
$aim-100-no-1\_6$&4&100&160\\
$aim-100-no-2\_0$&4&100&200\\
$aim-200-no-1\_6$&4&200&320\\
$aim-200-no-2\_0$&4&200&400\\
$uuf50-218$&100&50&218\\
\hline
\end{tabular}
\label{instances}
\end{center}
\caption{Characteristics of each set of problem instances}
\end{table}

We used the CPU time (in seconds) that each instance requires to be
solved by a given tool as a measure of performance, as well as the
number of iterations it takes to solve the instance. To better
illustrate the usefulness of each component added to {\sc Ranger},
each problem instance was run with different versions of {\sc Ranger}:
the original {\sc Ranger} code as a basis for comparisons (Original);
{\sc Ranger} with unit propagation look-ahead (UPLA); and {\sc Ranger}
with both unit propagation look-ahead and extended resolution
(UPLA+ER). Results for {\sc Gunsat} were also collected.

The results were obtained in an Intel Xeon 5160 server (3.0GHz,
1333Mhz, 4GB) running Red Hat Enterprise Linux WS 4. Each problem
instance was given a timeout of 1000 seconds. Results for each
instance were obtained from 10 runs using 10 different seeds.

Tables~\ref{percentage}, \ref{times} and~\ref{iterations} show the
results of, respectively, the percentage of instances solved, the CPU
time taken to solve each set of instances and the median number of
iterations performed by each solver (only taking into account the
successful runs).  The CPU time reported was computed as follows:
first, for each instance was considered the mean time of the
successful runs; second, the average of the given results for each
instance in a set of instances was computed.  The number of iterations
are not reported for {\sc Gunsat} given that such information is not
provided by the tool.

\begin{table}[t]
\begin{center}
\begin{tabular}{|c|cccc|}
\cline{2-5}
\multicolumn{1}{c|}{}&\multicolumn{4}{c|}{\% of instances solved}\\
\hline
Instance&Original&UPLA&UPLA+ER&{\sc Gunsat}\\
\hline
$aim-50-no-1\_6$& 100  & 100 & 100  & 100  \\
$aim-50-no-2\_0$& 75  & 100  & 100  & 100  \\
$aim-100-no-1\_6$& 100  & 100  & 100  & 97.5  \\
$aim-100-no-2\_0$& 100  & 100  & 100  & 100  \\
$aim-200-no-1\_6$& 75  & 77.5  & 55  & 80  \\
$aim-200-no-2\_0$& 55  & 66.2  & 37.5  & 80  \\
$uuf50-218$& 20 & 52.5  & 5.7 & 59.3  \\
\hline
\end{tabular}
\label{percentage}
\end{center}
\caption{Percentage of instances solved by {\sc Ranger} variants and {\sc Gunsat}}
\end{table}

\begin{table}[t]
\begin{center}
\begin{tabular}{|c|cccc|}
\cline{2-5}
\multicolumn{1}{c|}{}&\multicolumn{4}{c|}{Time (seconds)}\\
\hline
Instance&Original&UPLA&UPLA+ER&{\sc Gunsat}\\
\hline
$aim-50-no-1\_6$&0.015&0.096&0.281&1.21\\
$aim-50-no-2\_0$&250.026&0.114&0.384&1.41\\
$aim-100-no-1\_6$&0.834&0.91&13.57&36.96\\
$aim-100-no-2\_0$&11.35&1.74&24.09&3.27\\
$aim-200-no-1\_6$&259.227&24.56&317.1&16.72\\
$aim-200-no-2\_0$&512.156&238.7&818.2&379.8\\
$uuf50-218$&845.6&521.6&740.1&67.39\\
\hline
\end{tabular}
\label{times}
\end{center}
\caption{CPU time results for the {\sc Ranger} variants and {\sc Gunsat}}
\end{table}

\begin{table}[t]
\begin{center}
\begin{tabular}{|c|ccc|}
\cline{2-4}
\multicolumn{1}{c|}{}&\multicolumn{3}{c|}{Number of iterations}\\
\hline
Instance&Original&UPLA&UPLA+ER\\
\hline
$aim-50-no-1\_6$&39,223&4,201&4,958\\
$aim-50-no-2\_0$&67,105&6,240&7,026\\
$aim-100-no-1\_6$&433,398&134,572&120,122\\
$aim-100-no-2\_0$&3,266,805&160,971&174,131\\
$aim-200-no-1\_6$&5,902,875&1,390,580&1,788,612\\
$aim-200-no-2\_0$&9,802,581&3,066,713&3,762,236\\
$uuf50-218$&174,752,209&43,620,471&5,139,0176\\
\hline
\end{tabular}
\label{iterations}
\end{center}
\caption{Number of iterations for the {\sc Ranger} variants}
\end{table}

From these tables, we conclude that the best solver for solving easier
instances with few variables and a low ratio of variables/clauses is
the original {\sc Ranger} due to its simplicity. For all the other
instances of the {\tt aim} family, {\sc Ranger} with unit propagation
look-ahead beats the other two {\sc Ranger} variants in terms of
percentage of instances solved, required time and number of
iterations, and is comparable with {\sc Gunsat}. For the {\tt uuf50-218}
set of instances, {\sc Gunsat} is far superior to {\sc Ranger} in any
of its forms.

Not surprisingly, extended resolution proved to be too heavy for {\sc
Ranger}'s rather simple algorithm, and did not produce good
results. The strength of {\sc Ranger} lies in being simple enough to
perform many moves per second and that makes up for the rather simple
and somehow unintelligent but fast heuristics used. On the other hand,
extended resolution, while being a simple technique to implement, is
used in {\sc Gunsat} to improve its scoring scheme, which is a heavy
feature of the algorithm and is finely adjusted for optimum
performance. By implementing part of this scoring scheme in {\sc
Ranger} with the only goal of adding extended resolution to the
algorithm, we actually went against {\sc Ranger}'s principles of
simplicity, and thus the results achieved suggest not to use extended
resolution when {\sc Ranger} is concerned.

Thus we can conclude that adding simple and fast techniques to {\sc
Ranger} is a viable option when trying to improve its base
algorithm. These methods should not rely on scoring schemes, nor
depend too much on certain conditions to be met: they must be
independent and simple. This is based on the fact that they will be
added to {\sc Ranger} on its satisfiability-preserving transformation
phase, and will not alter the basic algorithm significantly.

\section {Conclusions and Future Work}

This paper evaluates the usefulness of integrating native techniques
to {\sc Gunsat} into {\sc Ranger}, another SLS solver able to prove
unsatisfiability.  We first tested both the original {\sc Ranger} and
{\sc Gunsat} in a number of unsatisfiable instance sets, in which {\sc
Gunsat} proved to be faster in the harder instances by systematically
beating {\sc Ranger}. This is mostly due to the fact that {\sc Gunsat}
has a more powerful reasoning mechanism and a finer heuristic to guide
moves, whereas {\sc Ranger} is simpler. Because of the use of powerful
high reasoning techniques like unit propagation look-ahead and
extended resolution, and a finer heuristic, {\sc Gunsat}'s moves are
slower but more intelligent and pondered, in contrast to {\sc
Ranger}'s rather faster but blinder moves.
 
The integration of {\sc Gunsat}'s techniques into {\sc Ranger}, namely
unit propagation look-ahead and extended resolution resulted in
improving {\sc Ranger}'s performance overall with the first technique,
whereas the second one has degraded {\sc Ranger}'s performance.

Future work includes further investigating why extended resolution 
impede the basic version of Ranger and using automatic methods to tune
parameters, namely $P_{er}$, such as the F-race proposed by Birattari 
{\em et al.}~\cite{birattari-gecco02}.

\subsubsection*{Acknowledgements}

This material is based in part upon works supported by the Science
Foundation Ireland under Grant No. 05/IN/I886, and partially
supported by Funda\-\c{c}\~{a}o para a Ci\^{e}ncia e Tecnologia under
research project PTDC/EIA/64164/2006. 

\bibliographystyle{eptcs}

\end{document}